\documentclass[runningheads]{llncs}
\usepackage{amsmath}
\usepackage[T1]{fontenc}
\usepackage{graphicx}
\usepackage{hyperref}
\usepackage{multirow}
\usepackage{booktabs}
\usepackage{makecell}
\usepackage{booktabs}
\usepackage{graphicx,verbatim}
\begin{document}
\title{Hybrid Explanation-Guided Learning for Transformer-Based Chest X-Ray Diagnosis}
\titlerunning{H-EGL}

\author{Shelley Zixin Shu \inst{1} \and Haozhe Luo \inst{1}\inst{5}\and Alexander Poellinger \inst{2}\inst{3} \and Mauricio Reyes\inst{1}\inst{4}}
\authorrunning{Shu et al.}

\institute{ARTORG Center for Biomedical Engineering Research, University of Bern, Murtenstrasse 50, Bern 3008, Switzerland \and Inselspital (Bern University Hospital), 3010 Bern, Switzerland \and Insel Gruppe Bern Universitätsinstitut für Diagnostische, Interventionelle und Pädiatrische Radiologie \and Department of Radiation Oncology, Inselspital, Bern University Hospital and University of Bern, Bern, Switzerland\and Kaiko.AI, Zurich, Switzerland \\}

\maketitle              
\begin{abstract}

Transformer-based deep learning models have demonstrated exceptional performance in medical imaging by leveraging attention mechanisms for feature representation and interpretability. However, these models are prone to learning spurious correlations, leading to biases and limited generalization. While human-AI attention alignment can mitigate these issues, it often depends on costly manual supervision.
In this work, we propose a Hybrid Explanation-Guided Learning (H-EGL) framework that combines self-supervised and human-guided constraints to enhance attention alignment and improve generalization. The self-supervised component of H-EGL leverages class-distinctive attention without relying on restrictive priors, promoting robustness and flexibility. We validate our approach on chest X-ray classification using the Vision Transformer (ViT), where H-EGL outperforms two state-of-the-art Explanation-Guided Learning (EGL) methods, demonstrating superior classification accuracy and generalization capability. Additionally, it produces attention maps that are better aligned with human expertise.

\end{abstract}

\section{Introduction}

Deep learning models, particularly transformer-based architectures, have demonstrated remarkable success across various medical image computing applications. One key enabler of this success is the attention mechanism \cite{dosovitskiy2020image}, which allows models to focus on the most relevant regions of an image, enabling superior feature representation and interpretability. However, despite their effectiveness, deep neural networks (DNNs) are inherently data-driven and prone to learn spurious correlations, leading to shortcut learning, biases, and fairness issues \cite{geirhos2020shortcut,degrave2021ai,gichoya2022ai}.

Human-AI alignment integrates human knowledge into model training to align attention with human-understood features, improving robustness and generalization \cite{rieger2020interpretations,wu2024gaze,wang2022follow}. This can involve expert-annotated explanations or iterative human feedback. However, the high cost of human annotations remains a major challenge. Human-AI attention alignment falls under Explanation-Guided Learning (EGL) \cite{gao2024going}, which uses explanations to guide learning. To reduce annotation needs, self-supervised EGL leverages intrinsic model constraints, but risks reinforcing spurious or misaligned explanations. Contrastive self-supervised explanation learning, such as \cite{pedapati2020learning}, lacks standardized methods for generating positive and negative samples and faces challenges in constructing such samples without ground truth \cite{gao2024going}. These methods often rely on rigid priors like sparsity, smoothness, and stability, which, while improving interpretability, may suppress subtle or complex feature, potentially missing critical clinical cues \cite{schramowski2020making}.

In response to these challenges, we propose a novel Hybrid Explanation-Guided Learning (H-EGL) approach that integrates both self-supervision and human supervision for attention alignment within the Explanation-Guided Learning (EGL) paradigm. This hybrid framework facilitates human-AI alignment while leveraging unlabeled human attention data, allowing self-supervision and human guidance to complement each other effectively. A key component of H-EGL is a self-supervised EGL method called Discriminative Attention Learning (DAL), which leverages class-distinctive attention maps from Vision Transformer (ViT) models. DAL was designed with inspiration from \cite{mahapatra2022interpretability}. Unlike prior self-supervised methods that impose rigid constraints, DAL introduces a flexible inductive bias that avoids over-regularization and preserves the model’s ability to learn complex, task-specific features. It promotes the distinctiveness of class-specific attention maps by guiding the model to generate discriminative attention outputs, leading to more robust and generalizable representations. Rather than relying on post-hoc interpretability tools, as in \cite{mahapatra2022interpretability} for convolutional networks, DAL directly exploits the inherent attention mechanisms of ViTs. To the best of our knowledge, this is the first application of a hybrid explanation-guided method within transformer architectures to assess and enhance human-AI alignment.

We evaluated our proposed method in the context of disease classification from chest X-ray images, a widely benchmarked medical imaging task. Using Vision Transformer (ViT) models, we compare our approach against two state-of-the-art EGL methods. To the best of our knowledge, this study represents the first evaluation of EGL on hybrid attention alignment for ViT models in medical imaging, which achieves superior classification performance and generates better expert-aligned attention maps. This emphasizes the complementary strengths of both paradigms.

\section{Methodology}

This section introduces Hybrid Explanation-Guided Learning (H-EGL), an approach that combines self-supervised and supervised EGL methods for Vision Transformer (ViT) models, applicable to a wide range of classification tasks. H-EGL consists of two key components: a self-supervised module called Discriminative Attention Learning (DAL), and a supervised module focused on human alignment. This paper demonstrate its capability in multi-label thoracic disease classification but notes that the approach is generic and can be extended to other medical image computing problems. We first present the overall framework, outlining its key components and their interactions. Then, we describe the methodology in detail, including the explanation guided learning strategy and attention alignment mechanism.

\begin{figure}[!htbp]
    \centering
    \includegraphics[width=0.9\textwidth]{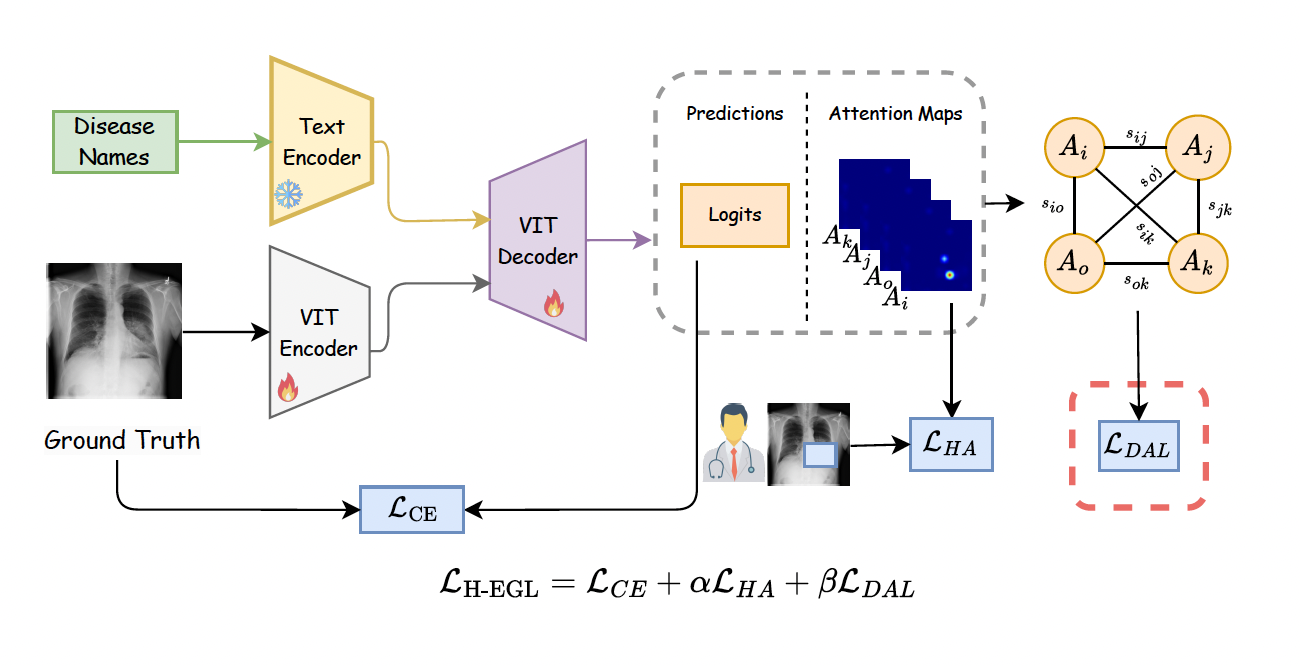}
\caption{
\textbf{Overview of the Hybrid Explanation-Guided Learning (H-EGL) framework.} H-EGL is a hybrid Explanation-Guided Learning (EGL) approach designed for Vision Transformers (ViTs). The model integrates visual inputs (e.g., chest X-rays) and textual information (e.g., disease labels such as \textit{Atelectasis}, \textit{Cardiomegaly}) via a ViT-based encoder-decoder architecture, employing a frozen text encoder and trainable ViT components. It generates both class-wise predictions and attention maps $A$. The H-EGL framework combines two key components: (i) a self-supervised module, \textbf{Discriminative Attention Learning (DAL)}, which encourages class-distinctive attention patterns, and (ii) a supervised \textbf{Human-AI Alignment} module. The total training objective incorporates:
(i) \textbf{Cross-Entropy Loss} $L_{CE}$ for classification,
(ii) \textbf{Human-AI Alignment Loss} $L_{HA}$ to align attention maps with expert annotations (e.g., bounding boxes), and (iii) \textbf{Discriminative Attention Loss} $L_{DAL}$, which enforces separation among class-specific attention maps. The figure illustrates attention distinctiveness (e.g., similarity $s_{ij}$ between attention maps $A_i$ and $A_j$) in a four-class setting, though the approach generalizes to any number of classes.}
\label{fig:overall_chart}
\end{figure}
We propose a unified framework, illustrated in Fig.~\ref{fig:overall_chart}, called Hybrid Explanati-on-Guided Learning (H-EGL), a novel method that integrates both human-supervised and self-supervised explanation-guided learning (EGL) for Vision Transformer (ViT) models. It is built upon DWARF \cite{luo2024dwarf}, a ViT-based encoder-decoder architecture tailored for disease classification from chest X-rays. DWARF fuses visual inputs (e.g., chest X-rays) and textual inputs (e.g., disease names like \textit{Atelectasis}, \textit{Cardiomegaly}) via cross-attention and aligns the model’s attention maps with human-provided annotations. In this architecture, the text encoder is frozen, while the ViT encoder and decoder are trainable, allowing dynamic refinement of visual features while preserving semantic consistency from the textual side. The model outputs multi-label disease classification predictions and class-specific attention maps denoted as \( A \).

H-EGL is designed to improve model interpretability and classification performance by jointly leveraging expert annotations and unlabeled data. It combines two complementary components: (i) a \textbf{human-AI alignment} loss that encourages the model to focus on expert-defined regions of interest, and (ii) a self-supervised module called \textbf{Discriminative Attention Learning (DAL)}, which promotes class-distinctive attention behavior across categories. The H-EGL framework is optimized using a composite loss function:

\begin{equation}
    \mathcal{L}_{\text{H-EGL}} = \mathcal{L}_{\text{CE}} + \alpha \mathcal{L}_{\text{HA}} + \beta \mathcal{L}_{\text{DAL}}.
    \label{H-EGL eq}
\end{equation}

The weights \(\alpha > 0\) and \(\beta > 0\) control the relative contribution of the two explanation components. This formulation generalizes to both labeled and unlabeled human attention data and can be easily extended to other transformer-based classification tasks. \(\mathcal{L}_{\text{CE}}\) is the standard multi-label cross-entropy loss for classification. \(\mathcal{L}_{\text{HA}}\) is the human-AI alignment loss, aligns attention maps with expert-annotated pathology regions, adopted a penalized Dice loss following \cite{luo2024dwarf}:

\begin{equation}
    \mathcal{L}_{\text{HA}} = 1 - \frac{2 \times |A_i \odot M_i|}{|A_i| + |M_i| + w_{FP} N_{FP}},
\end{equation}

where \( A_i \) is the model-generated attention map for class \( i \), \( M_i \) is the corresponding expert mask, \( \odot \) denotes pixel-wise multiplication, \( N_{FP} \) is the number of false positives, and \( w_{FP} \) is a penalty coefficient. \(\mathcal{L}_{\text{DAL}}\) is the proposed discriminative attention learning loss, enforces class-level separability in the attention maps by minimizing their pairwise similarity, 
\begin{equation}
 \mathcal{L}_{\text{DAL}} = \frac{2}{C(C-1)} \sum_{i=1}^{C-1} \sum_{j=i+1}^{C} \left| S(A_i, A_j) \right|,
\end{equation}

where \( C \) is the number of classes, and \( S(A_i, A_j) \) denotes the cosine similarity between attention maps \( A_i \) and \( A_j \):

\begin{equation}
    S(A_i, A_j) = \frac{\langle \mathbf{A}_i, \mathbf{A}_j \rangle}{\|\mathbf{A}_i\| \, \|\mathbf{A}_j\|}.
\end{equation}

This self-supervised mechanism encourages the model to produce attention maps that are more discriminative across different classes. Unlike traditional contrastive methods, DAL avoids the need for negative sample generation or image perturbation, making it efficient and scalable.

In summary, H-EGL integrates human-supervised and self-supervised attention alignment in a unified framework, improving both interpretability and performance. It supports learning from expert-labeled data while also leveraging large-scale unlabeled datasets through a lightweight, task-relevant inductive bias. This design offers a flexible and generalizable pathway toward robust human-AI aligned vision models.

\section{Experiments}

We evaluated H-EGL on the task of classifying four prevalent thoracic pathologies, atelectasis, cardiomegaly, consolidation, and effusion, using chest X-ray images. These conditions are well-represented in public datasets providing human-annotated attention maps, making them ideal for benchmarking \cite{luo2024dwarf}. We evaluate the model performance with various classification metrics and also did an ablation study by removing the individual component of the hybrid method to evaluation the individual contribution of DAL and human-AI alignment. Additionally, to evaluate model robustness and generalization, we measured the generalization gap, defined as the difference between the performance on validation and test sets \cite{d2022underspecification}. We applied Gaussian noise with mean zero and various standard deviations to further examine the robustness of the model again noise at test time. 

\subsubsection{Dataset:}
We utilized the ChestXDet dataset \cite{liu2020chestx}, a subset of NIH ChestX-ray14 \cite{wang2017chestx}, including human-annotated pathology location segmentation. ChestXDet consists of 3,578 patients, with 3,025 samples in the training set and 553 in the test set. This official train-test split was used in our experiments. Each image includes pathology annotations (marked using bounding boxes and polygons), verified by three radiologists. We used an 80-20 train-validation split. All evaluations are conducted on the test set. 

\subsubsection{Implementation:}
We conducted experiments by training the model five times, each with a different random seed used to split the training and validation sets from the official training data. The models were then evaluated on the separate official test set. The reported results represent the average performance across these five runs on the official test set. We use Med-KEBERT as text encoder and Transformer Query Network decoder for cross-attention (both as part of KAD \cite{zhang2023knowledge}). 
A ViT-B \cite{dosovitskiy2020image} with a 224×224 input resolution  is used as an image encoder. The models achieving the highest AUC on the validation set were used for evaluation. The attention maps are obtained from the decoder's cross-attention layers. Optimization was performed using the AdamW optimizer with a learning rate of 1e-5. Models were trained for 1000 epochs with early stopping, patience set to 50 and a 20-epoch warm-up phase. The training used a batch size of 32. The penalty dice score weight $w_{FP}$ is 1, and hyperparameters $\alpha$ and $\beta$ were set to $1.0$ in all experiments when the $\mathcal{L}_{HA}$ or $\mathcal{L}_{DAL}$ module was added for H-EGL. All experiments run on RTX 4090 GPUs with CUDA v12.2.

\subsubsection{Baselines:}
We evaluated two strong baseline models in our experiments. KAD \cite{zhang2023knowledge} is a knowledge-aware detection framework that enhances visual reasoning by incorporating human-based knowledge graphs and structured knowledge (e.g., RadGraph \cite{jain2021radgraph}) to better capture complex visual relationships. It serves as the foundational backbone of our model. GAIN \cite{li2018tell} is a gradient-based attention network aimed at improving interpretable visual recognition through refined attention mechanisms. For our baseline, we adapted GAIN's loss function to work with a transformer backbone, applying its gradient-based attention strategy for EGL. We compare the performance of these baselines with our Hybrid Explanation-Guided Learning (H-EGL) method, which integrates self-supervision and human annotations for EGL. Additionally, we assess the standalone performance of DAL as a purely self-supervised EGL approach ($\alpha=0$), and DWARF \cite{luo2024dwarf} ($\beta=0$), which relies solely on human-annotation-guided explanation learning.

\section{Results} 

We evaluated model performance using classification metrics including AUC, F\(_1\), and MCC. To assess generalization ability, we also measured the performance gap between validation and test sets, following the methodology in~\cite{d2022underspecification}.

Table~\ref{performance table} presents a comparative evaluation of H-EGL against other state-of-the-art methods. The proposed H-EGL achieved the highest test AUC (89.3\%), outperforming KAD (88.1\%) and the prior method GAIN (88.0\%), while also exhibiting a low variance at 0.7\%. Notably, H-EGL further improved classification performance, achieving the highest \textbf{F\(1\)\(_{\text{test}}\)} (69.4\%) and MCC\(_{\text{test}}\) (58.3\%), while significantly reducing F\(1\)\(_{\text{gap}}\) (0.5\%) and MCC\(_{\text{gap}}\) (3.8\%). These results indicate enhanced robustness and consistency across test samples. 

Moreover, H-EGL outperformed both purely self-supervised (AUC 89.3$\pm1.0\%$) and human-annotation-only methods (AUC 88.4\%$\pm 0.2\%$) for explanation-guided learning (EGL). The self-supervised EGL ($\alpha$=0) also demonstrated competitive performance relative to DWARF ($\beta$=0), despite relying solely on self-supervised signals. These findings underscore that \textit{H-EGL} achieves the best trade-off between performance and stability—reinforcing the effectiveness of combining self-supervised and human-guided attention constraints.

\begin{table}[!htb]
    \centering
    \caption{
  \textbf{Performance and ablation comparison of H-EGL and baseline methods.}
This table compares the proposed H-EGL model against baseline methods: KAD~\cite{zhang2023knowledge}, which utilizes knowledge graphs for improved visual reasoning, and GAIN~\cite{li2018tell}, which enhances interpretability via attention guided by cross-entropy loss. An ablation study is conducted by removing key components of H-EGL: the human-annotated explanation loss ($L_{HA}$) and the self-supervised EGL loss ($L_{DAL}$). $\alpha$ and $\beta$ refer to Eq. \ref{H-EGL eq}. Results are reported as mean~$\pm$~standard deviation over five runs. 
}

    \begin{tabular}{lcccccc}

    \toprule
    & AUC$_{test}\uparrow$  & AUC$_{gap}\downarrow$ & F1$_{test}\uparrow$ & F1$_{gap}\downarrow$ & MCC$_{test}\uparrow$ & MCC$_{gap}\downarrow$\\
    \midrule
    KAD \cite{zhang2023knowledge} & 88.1$\pm$0.3\% &2.5\% &68.2$\pm$2.5\%&1.8\%&  57.5$\pm$2.3\%& 4.8\%\\
    GAIN \cite{li2018tell} & 88.0$\pm$0.4\%&2.7\%& 67.8$\pm$2.2\%&2.4\%& 57.2$\pm$2.0\%&5.6\%\\
    H-EGL (Ours) & \textbf{89.3$\pm$0.7\%} & 1.5\% & \textbf{69.4$\pm$1.9\%} & \textbf{0.5\%} & \textbf{58.3$\pm$2.5\%} & \textbf{3.8\%} \\
    \midrule
    \multirow{2}{*}{\makecell[l]{H-EGL\\ $\alpha$=0}} 
        & \multirow{2}{*}{\textbf{89.3$\pm$1.0\%}} 
        & \multirow{2}{*}{\textbf{1.4\%}} 
        & \multirow{2}{*}{67.6$\pm$1.2\%} 
        & \multirow{2}{*}{1.4\%} 
        & \multirow{2}{*}{56.5$\pm$1.6\%} 
        & \multirow{2}{*}{5.2\%} \\
    \\
    $\beta$=0 \cite{luo2024dwarf}  & 88.4$\pm$0.2\% & 2.5\% & 66.9$\pm$1.2\% & 3.2\% & 56.3$\pm$1.0\% & 6.5\%\\
    \bottomrule
\end{tabular}
    \label{performance table}
\end{table}

\begin{figure}[!htb]
    \centering
    \includegraphics[width=0.7\textwidth, height=5cm, keepaspectratio]{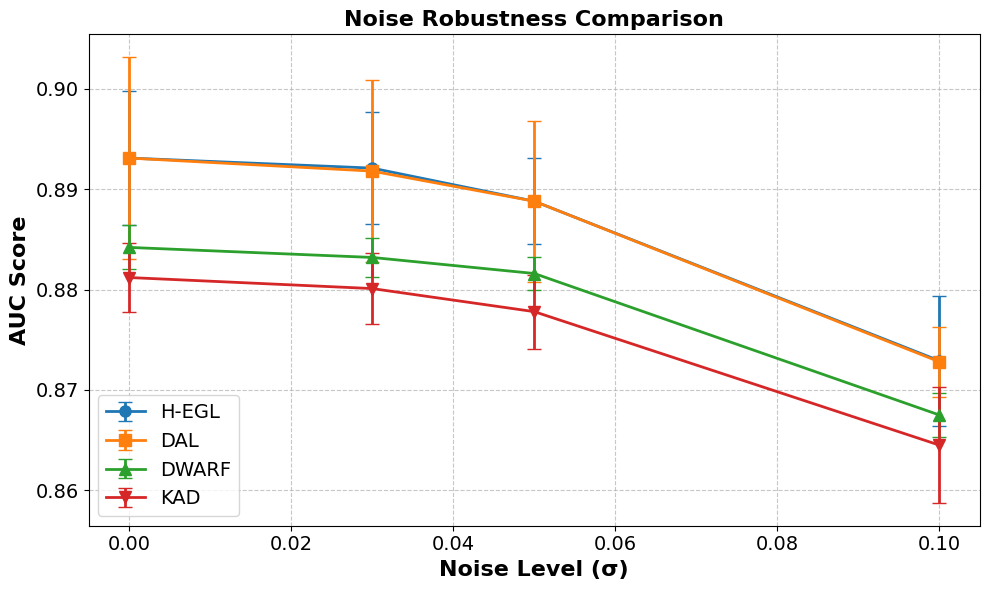}
    \caption{Robustness of models under different levels of perturbed inputs. The noises applied are normally distributed with mean zero and different standard deviation levels ($\sigma=0, 0.03, 0.05, 0.1$). Note: H-EGL and DAL curves closely overlap.}
    \label{fig:noise_comparision}
\end{figure}

Figure \ref{fig:noise_comparision} shows the results of a robustness analysis of models when confronted with different levels of noise added to test images. Overall, while all models show a decrease in performance at increased levels of perturbations, the proposed H-EGL approaches remain superior to the other benchmarked approaches.

\textbf{Qualitative Attention Results:}
\begin{figure}[!htb]
    \centering
    \includegraphics[width=0.9\textwidth]{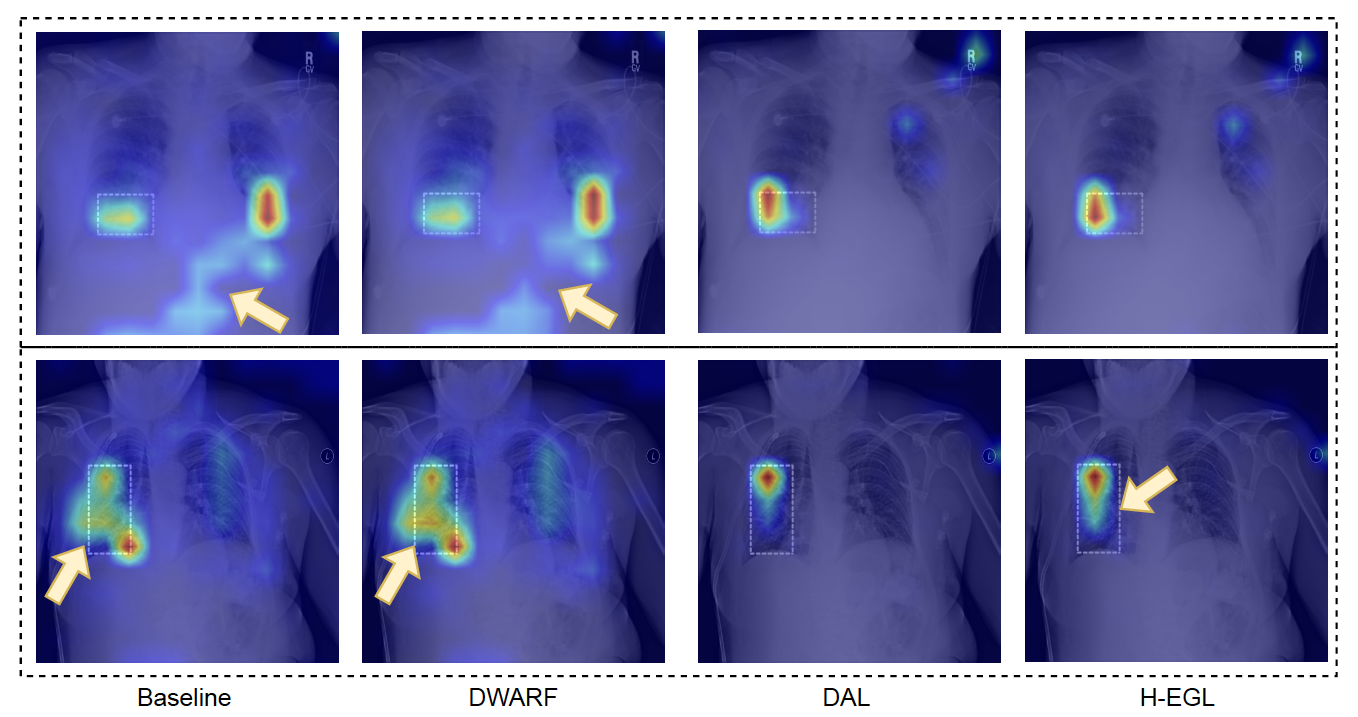}
    \caption{Comparison of attention maps from the baseline model ($\alpha=0$, $\beta$=0), DWARF ($\beta=0$), DAL ($\alpha=0$), and H-EGL model. Abnormal pathologies are highlighted with white bounding boxes, while arrows indicate areas with visible changes across models. }
    \label{fig:attention_maps}
\end{figure}
In addition to quantitative results, we visualized the attention maps of the baseline KAD model, DWARF, the proposed self-supervised approach (DAL), and the proposed Hybrid EGL approach (H-EGL). As shown in Fig.~\ref{fig:attention_maps}, the baseline KAD model effectively coincides with the human annotation (white dotted outline) but also wrongly highlights the lower lobe of both lungs. The DWARF model reduces false positives in the lower part of the image (see yellow arrow) but also wrongly focuses on the left lung. In contrast, H-EGL and DAL models more accurately identify the pathological regions while significantly reducing the false positives.

\section{Discussion and Further Work}
We introduce a Hybrid Explanation-Guided Learning (H-EGL) approach that combines self-supervision and human annotations to improve performance and generate attention maps better aligned with expert knowledge. The self-supervised component introduces an inductive bias that encourages the model to learn distinctive, class-specific attention patterns. H-EGL is architecture-agnostic and flexible. Its self-supervised component DAL imposing no constraints on feature selection, requiring no localization annotations, and eliminating the need for explicitly defined negative samples.

Our experiments show that incorporating discriminative attention guidance significantly boosts AUC scores, with performance dropping noticeably when this component is removed. H-EGL also demonstrates strong robustness under noisy test conditions, consistently outperforming baseline methods. By leveraging both labeled and unlabeled data, the framework improves attention quality and classification accuracy while remaining scalable, annotation-efficient, and compatible with existing transformer-based medical imaging models.

A key insight from our findings is the need to balance human alignment and self-supervision for explanation-guided learning. While human-guided alignment steers the model toward clinically meaningful features, fully supervised methods can be costly and may result in rigid attention behaviors that lack generalization. On the other hand, purely self-supervised strategies, though scalable, risk encouraging shortcut learning when unconstrained. H-EGL addresses this by integrating both paradigms, enabling human-guided supervision to refine model explanations without overly restricting its learning capacity. This approach parallels findings in policy learning, where Trust Region (TR) methods prevent overoptimization by adaptively adjusting constraints during training \cite{gorbatovski2024learn}. Similarly, H-EGL strikes a balance between autonomy and guidance, promoting both robustness and generalization.

We further observe that DWARF faces a trade-off between interpretability and performance: strong human alignment can reduce classification accuracy (see Table \ref{performance table} and Fig. \ref{fig:attention_maps}). In contrast, H-EGL effectively localize pathological regions without compromising accuracy, achieving a better balance between interpretability and predictive power. Inspired by policy learning strategies, future work will explore dynamic mechanisms to optimize the degree of self-supervision and human alignment during training, further enhancing interpretability, generalization, and robustness in medical image analysis.

\section{Conclusion}

We propose a hybrid attention alignment approach to improve the robustness and interpretability of transformer-based medical imaging models. The Hybrid Explanation-Guided Learning (H-EGL) framework combines human-guided and self-supervised learning to localize pathological regions and boost classification performance. Results validate H-EGL as a scalable alternative to fully supervised alignment. Future work will extend to larger datasets and explore dynamic alignment strategies to further enhance interpretability and generalization in clinical tasks.

\section*{Acknowledgments}
We acknowledge funding by the Swiss National Science Foundation (project number 212939). We report no financial relationship or conflicts of interest.

\bibliographystyle{splncs04}
\bibliography{reference}

\end{document}